\documentclass{article}

\usepackage{arxiv}

\usepackage[utf8]{inputenc} % allow utf-8 input
\usepackage[T1]{fontenc}    % use 8-bit T1 fonts
\usepackage{hyperref}       % hyperlinks
\usepackage{url}            % simple URL typesetting
\usepackage{booktabs}       % professional-quality tables
\usepackage{amsfonts}       % blackboard math symbols
\usepackage{nicefrac}       % compact symbols for 1/2, etc.
\usepackage{microtype}      % microtypography
\usepackage{lipsum}
\usepackage{graphicx}
\graphicspath{ {./images/} }

\usepackage{microtype}
\usepackage{times}
\usepackage{soul}
\usepackage{url}
\usepackage[utf8]{inputenc}
\usepackage[small]{caption}
\usepackage{graphicx}
\usepackage{amsmath}
\usepackage{amsthm}
\usepackage{booktabs}
\usepackage{algorithm}
\usepackage{algorithmic}
\usepackage[shortlabels]{enumitem}
\usepackage[TABBOTCAP]{subfigure}
\usepackage{multirow}
\usepackage{diagbox}

\def\CM{{\mathcal C}}

\def\IM{{\mathcal I}}

\def\SM{{\mathcal S}}

\def\0{{\bf 0}}
\def\1{{\bf 1}}

\usepackage{xcolor}

\title{A Brief Survey and Comparative Study of Recent Development of Pronoun Coreference Resolution}

\author{
  Hongming Zhang, Xinran Zhao, Yangqiu Song\\
  Department of Computer Science, HKUST\\
  \texttt{hzhangal@cse.ust.hk, xzhaoar@connect.ust.hk, yqsong@cse.ust.hk} \\
}

\begin{document}
\maketitle
\begin{abstract}
Pronoun Coreference Resolution (PCR) is the task of resolving pronominal expressions to all mentions they refer to. 
Compared with the general coreference resolution task, the main challenge of PCR is the coreference relation prediction rather than the mention detection.
As one important natural language understanding (NLU) component, pronoun resolution is crucial for many downstream tasks and still challenging for existing models, which motivates us to survey existing approaches and think about how to do better.
In this survey, we first introduce representative datasets and models for the ordinary pronoun coreference resolution task.
Then we focus on recent progress on hard pronoun coreference resolution problems (e.g., Winograd Schema Challenge) to analyze how well current models can understand commonsense.
We conduct extensive experiments to show that even though current models are achieving good performance on the standard evaluation set, they are still not ready to be used in real applications (e.g., all SOTA models struggle on correctly resolving pronouns to infrequent objects). 
All experiment codes are available at: \url{https://github.com/HKUST-KnowComp/PCR}.
\end{abstract}

% keywords can be removed

\section{Introduction}
\label{sec:introduction}

The question of how human beings resolve pronouns\footnote{Some pronouns may refer to non-nominal antecedents. For example, the pronoun ``it'' in ``It is too cold in the Winter here'' does not refer to any real object~\cite{kolhatkar-etal-2018-survey}. But in this survey, we only focus on pronouns that refer to nominal antecedents.} has long been of interest to both linguistic and natural language processing (NLP) communities, for the reason that a pronoun itself only having weak semantic meaning brings challenges to natural language understanding.
To explore solutions for that question, pronoun coreference resolution (PCR)~\cite{hobbs1978resolving} was proposed.\footnote{Previous studies~\cite{ng2005supervised,DBLP:conf/acl/ZhangSSY19} mainly focus on three kinds of pronouns: third personal pronoun (e.g., \textit{she}, \textit{her}, \textit{he}, \textit{him}, \textit{them}, \textit{they}, \textit{it}), possessive pronoun (e.g., \textit{his}, \textit{hers}, \textit{its}, \textit{their}, \textit{theirs}), and demonstrative pronoun (e.g., \textit{this}, \textit{that}, \textit{these}, \textit{those}).
The first and second personal pronouns are typically not considered as they often refer to the current speakers, which are normally out of the conversation or document.
Besides that, conventional PCR works~\cite{ng2005supervised,DBLP:conf/acl/ZhangSSY19,DBLP:conf/naacl/ZhangSS19} mostly focusing on identifying coreference relations between pronouns and noun phrases rather than coreference relation between pronouns.}
As a challenging yet vital natural language understanding task, pronoun coreference resolution is to find the correct reference for a given pronominal anaphor in the context and has been shown to be crucial for a series of downstream tasks, such as machine translation~\cite{mitkov1995anaphora}, summarization~\cite{steinberger2007two}, and dialog systems~\cite{strube2003machine}.

To investigate the difference between PCR and the general coreference resolution task, which tries to identify not only the coreference relations between noun phrases (NP) and pronouns (P) but also potential coreference relations between noun phrases or coreference relations between pronouns, we conduct experiments with one recent breakthrough model (i.e., End-to-end model~\cite{DBLP:conf/emnlp/LeeHLZ17}) on the CoNLL-2012 shard task~\cite{DBLP:conf/conll/PradhanMXUZ12} under two settings: one without the gold mention and one with the gold mention.
In the `without gold mention' setting, models are required to first identify spans from the documents as the mentions and then predict the coreference relations among these mentions.
As a comparison, if gold mentions are provided, models only need to predict the coreference relations.
From the results in Table~\ref{tab:PCR-vs-generalCR} we can see that, without gold mention, the model performs well on P-P coreference relations while not that well on the other two kinds of relations. However, if gold mentions are provided, the model can achieve very good performance on the NP-NP coreference relations.
Compare with other kinds of coreference relations, no matter whether the gold mention is provided or not, resolving pronouns to noun phrases is always the most challenging one. 
% the main challenge of finding coreference relations between noun phrases is probably \revisexr{the} mention detection rather than the coreference prediction due to the dataset distribution of CoNLL-2012 (e.g., most noun phrases that coref 
% to each other are very similar). Different from them, no matter whether the gold mention is provided or not, resolving pronouns to noun phrases is always the most challenging one. 

\begin{table}[t]
\small
    \centering
    \begin{tabular}{c||c|c|c}
    \toprule
    Type & \# Pairs & F1 (no mention) & F1 (mention)\\
         \midrule
         NP-NP & 25,828 & 0.690 & 0.768 \\
         NP-P & 43,883 & 0.667 & 0.707 \\
         P-P & 41,741 & 0.754 & 0.763 \\
         \midrule
         Overall & 111,452 & 0.705 & 0.742 \\
    \bottomrule
    \end{tabular}
    \vspace{0.1in}
    \caption{The performance of the End-to-end model on the CoNLL-2012 shared task coreference resolution dataset. The model's performances of different coreference types are reported separately. }
    \label{tab:PCR-vs-generalCR}
\end{table}

The correct resolution of pronouns typically requires reasoning over both linguistic knowledge (e.g., `they' typically can only refer to plural objects\footnote{The only exception is the organisation named entities. For example, ``they'' can refer to ``the company''~\cite{hardmeier-etal-2018-forms}.}) and commonsense knowledge (e.g., in sentence ``The fish ate the worm, it was hungry'', `it' refers to `fish' because hungry things tend to eat rather than being eaten.).
Considering that the ordinary PCR task evaluates the inference over both types of knowledge at the same time, the performance on ordinary PCR tasks cannot clearly reflect models' performance regarding different knowledge types.
To address this problem, the Winograd Schema Challenge (WSC)~\cite{levesque2012winograd} task is proposed. The influence of all commonly used linguistic knowledge is avoided during the creation of WSC such that WSC can be used to reflect how current PCR models can understand commonsense knowledge.
In Section~\ref{sec:OrdinaryPCR} and~\ref{sec:HardPCR}, we introduce the progress and remaining challenges on the ordinary PCR and WSC tasks respectively.
After that, we introduce other PCR tasks that are developed for different research purposes in Section~\ref{sec:OtherPCR}.
In the end, we conclude this survey with Section~\ref{sec:conclusion}.
The contribution of this survey is three-fold: (1) we broadly introduce available PCR tasks, datasets, and models; (2) We summarize the main contribution of recent models; (3) We conduct experiments to analyze the limitations of current models, which can help the community think about how to better solve PCR in the future.

\section{Ordinary PCR}\label{sec:OrdinaryPCR}

Ordinary pronoun coreference resolution tasks are often defined over formal textual corpus (e.g., newspaper) and the 
% For example, ACE~\cite{DBLP:conf/lrec/DoddingtonMPRSW04} is developed on newspapers and newswire corpus.
annotation is usually conducted by domain experts or linguists.
The PCR task can be formally defined as follows.
Given a text $D$, which contains a pronoun $p$, the goal is to identify all the mentions that $p$ refers to.
We denote the correct mentions $p$ refers to as $c \in \CM$, where $\CM$ is the correct mention set.
Similarly, each candidate span is denoted as $s \in \SM$, where $\SM$ is the set of all candidate spans.
Note that in the case where no golden mentions are provided, all possible spans in $D$ are used to form $\SM$.
The task is thus to identify $\CM$ out of $\SM$.
In the rest of this section, we introduce the widely used datasets as well as the progress and limitation of current approaches.

\subsection{Datasets}

Throughout the years, researchers in the NLP community have devoted great efforts to developing high-quality coreference resolution datasets\footnote{Some datasets (e.g., CoNLL-2012 shared task) are originally designed for the general coreference resolution task. Nonetheless, we can easily convert them into a PCR task.} and we introduce representative ones as follows:

\begin{enumerate}[leftmargin=*]
    \item \textbf{MUC}: MUC-6~\cite{DBLP:conf/coling/GrishmanS96} and MUC-7~\cite{chinchor1998overview}, which were developed for the 6$^{th}$ and 7$^{th}$ message understanding conferences respectively, are the earliest coreference resolution datasets. They are focusing on English news articles and are relatively small compared with modern datasets.
    \item \textbf{ACE}: The ACE dataset~\cite{DBLP:conf/lrec/DoddingtonMPRSW04} was proposed as part of the Automatic Content Extraction program.
    Compared with MUC datasets, ACE extends the corpus domain from news to other domains like telephonic speeches and broadcast conversations.
    \item \textbf{CoNLL shared tasks}: CoNLL-2011~\cite{DBLP:conf/conll/PradhanRMPWX11} and CoNLL-2012~\cite{DBLP:conf/conll/PradhanMXUZ12} shared tasks were proposed to evaluate models' abilities of resolving unrestricted coreference resolution.
    Among these two, CoNLL-2011 only contains annotation about English and CoNLL-2012 extends to multilingual (e.g., Chinese and Arabic).
    Compared with MUC and ACE, CoNLL shared tasks have a much larger scale.
    Moreover, as CoNLL-2012 shared tasks provide clear training, dev, and test set separation as well as the official evaluation tool, it is the most widely used evaluation benchmark for the coreference resolution task.
    \item \textbf{WikiCoref}: Recently, a new coreference dataset WikiCoref~\cite{DBLP:conf/lrec/GhaddarL16} was proposed as a supplementary of CoNLL shared tasks. Different from CoNLL, where most of the corpus is from the newswire, WikiCoref directly annotates Wikipedia pages, which provides a new way to evaluate models' performances in the out-of-domain setting. 
    \item \textbf{Crowd-sourced Coref}: \cite{DBLP:conf/naacl/PoesioCPYUK19} leveraged a crowd-sourced game to collect 2.2 million annotations about 108,000 coreference relations, which makes it one of the largest coreference dataset. Moreover, their annotations also include ambiguous coreference relations.
\end{enumerate}

\subsection{Methods}

In this subsection, we introduce representative models for the ordinary PCR task. We first briefly introduce conventional approaches that rely on human-designed rules or features and then introduce the end-to-end model, which is a groundbreaking model for solving coreference resolution tasks. After that, we briefly introduce a few recent improvements over the end-to-end model.

\subsubsection{Rule and Feature Based Methods}
Before the deep learning era, human-designed rules~\cite{hobbs1978resolving,DBLP:conf/emnlp/RaghunathanLRCSJM10}, knowledge~\cite{ponzetto-strube-2006-exploiting,Versley2016}, or features~\cite{ng2005supervised,wiseman-etal-2016-learning} dominated the general coreference resolution and PCR tasks. Some rules and features are crucial for correctly resolving pronouns~\cite{LeeCPCSJ13}. For example, `he' can only refer to males and `she' can only refer to females; `it' can only refer to singular objects and `them' can only refer to plural objects.
The performances of these methods heavily rely on the coverage and quality of the manually defined rules and features.
Based on these designed features~\cite{BengtsonR08}, a few more advanced machine learning models were applied to the coreference resolution task. For example, instead of identifying coreference relation pair-wisely, \cite{DBLP:conf/acl/ClarkM15} proposes an entity-centric coreference system that can learn an effective policy for building coreference chains incrementally. Besides that, a novel model was also proposed to predict coreference relations with a deep reinforcement learning framework~\cite{DBLP:conf/emnlp/ClarkM16}. 
%Both models once achieved the state-of-the-art performance on the general coreference resolution task.
Moreover, heuristic rules based on linguistic knowledge can also be incorporated into constraints for machine learning models~\cite{ChangSR13}.
% However, as their original paper did not analyze how good they can resolve pronouns, we could not directly compare them with older models designed specifically for the PCR task~\cite{ng2005supervised}.

\subsubsection{End-to-end Model}
% \revisexr{do we need more elaboration on how this kind of framework work? how they capture the contextual information?}
Leveraging human-designed rules or features can help accurately resolve some pronouns, but it is hard to manually design rules to cover all cases. 
To solve this problem, an end-to-end deep model~\cite{DBLP:conf/emnlp/LeeHLZ17} was proposed.
Different from other machine learning-based methods, it does not use any human-defined rules, yet achieves surprisingly good performance.
Specifically, the end-to-end model first leverages the combination of Bi-directional LSTM and inner-attention modules to encode local context and generate representations for all potential mentions.
After that, a standard feed-forward neural network is used to predict the coreference relations.
Experiment results show that the proposed model is simple yet effective.
Its success proves that current deep models are capable of capturing rich contextual information, which is crucial for resolving coreference relations. 

\begin{table*}[t]
\small
    \centering
    \begin{tabular}{l|ccc|ccc|ccc|ccc}
        \toprule
         \multirow{ 3}{*}{Model} & \multicolumn{3}{c|}{Third Personal} & \multicolumn{3}{|c|}{Possessive} & \multicolumn{3}{c|}{Demonstrative} &  \multicolumn{3}{|c}{Overall} \\
         & \multicolumn{3}{c|}{(18,147)} & \multicolumn{3}{|c|}{(6,843)} & \multicolumn{3}{c|}{(546)} &  \multicolumn{3}{|c}{(25,536)}
         \\
         & P & R & F1 & P & R & F1 & P & R & F1 & P & R & F1 \\
         
         \midrule
         Deterministic~\tiny{\cite{DBLP:conf/emnlp/RaghunathanLRCSJM10}} & 25.5 & 58.9 & 35.6 & 22.9 & 64.3 & 33.8 & 3.4 & 5.7 & 4.2 & 23.4 & 57.0 & 33.4  \\ 
         Statistical~\tiny{\cite{DBLP:conf/acl/ClarkM15}} & 25.8 & 62.1 & 36.5 & 28.9 & 64.9 & 40.0 & 9.8 & 6.3 & 7.6 & 25.4 & 59.3 & 36.5 \\ 
         Deep-RL~\tiny{\cite{DBLP:conf/emnlp/ClarkM16}} & 78.6 & 63.9 & 70.5 & 73.3 & 68.9 & 71.0 & 3.7 & 2.9 & 5.5 & 76.4 & 61.2 & 68.0 \\ 
         \midrule
         End-to-end~\tiny{\cite{DBLP:conf/emnlp/LeeHLZ17}} & 70.7 & 77.8 & 74.1 & 75.6 & 74.0 & 74.8 & 37.8 & 71.7 & 49.5 & 68.3 & 76.4 & 72.1 \\
         \midrule 
        %  Without KG & 78.2 & 72.4 & 75.2 & 80.0 & 66.4 & 72.6 & 46.7 & 62.5 & 53.4 & 75.7 & 70.1 & 72.8 \\
        %  Without Attention & 76.6 & 77.9 & 77.2 & 79.0 & 73.5 & 76.2 & 42.4 & 72.6 & 53.5 & 73.6 & 76.4 & 74.9 \\
         $\quad$ + KG~\tiny{\cite{DBLP:conf/acl/ZhangSSY19}} & 80.0 & 75.6 & 77.7 & 81.7 & 72.2 & 76.7 & 50.8 & \textbf{64.6} & \textbf{56.9} & 77.9 & 74.0 & 75.9 \\
         $\quad$ + SpanBERT~\tiny{\cite{DBLP:journals/tacl/JoshiCLWZL20}} &\textbf{82.4}  &\textbf{80.5}  & \textbf{81.5}  &\textbf{83.9}  &\textbf{81.0} &\textbf{82.4} &\textbf{52.0} &61.5 &56.4  &\textbf{82.2} &\textbf{80.2} &\textbf{81.2} \\
         \bottomrule
    \end{tabular}
    \vspace{0.1in}
    \caption{Performances of different models on the CoNLL-2012 shared task. Precision (P), recall (R), and the F1 score are reported. Numbers of different types of pronouns in the test set are shown in the brackets. Best models are indicated with the \textbf{bold} font.}
    \label{tab:ordinary-main_result}
\end{table*}
\subsubsection{Further Improvements}

Recently, on top of the end-to-end model, a few improvement works were proposed to address different limitations of the original end-to-end model\footnote{These models once achieved better performance either on the general coreference resolution task or the PCR task. }:
\begin{enumerate}[leftmargin=*]
    \item \textbf{Higher-order Information}: One limitation of the original end-to-end model is that all predictions are based on pairs, which is not sufficient for capturing higher-order coreference relations. To fix this issue, a differentiable approximation module was proposed in ~\cite{DBLP:conf/naacl/LeeHZ18} to provide the higher-order coreference resolution inference ability (i.e., leveraging the coreference cluster to better predict the coreference relations).
    Moreover, this work first incorporates ELMo~\cite{DBLP:conf/naacl/PetersNIGCLZ18} as part of the word representation, which is proven very effective.
    \item \textbf{Structured Knowledge}: Another limitation of the end-to-end model is that its success heavily relies on the quality and coverage of the training data.
    However, in real applications, it is labor-intensive and almost impossible to annotate a large-scale dataset to contain all scenarios. 
    To solve this problem, two works~\cite{DBLP:conf/naacl/ZhangSS19,DBLP:conf/acl/ZhangSSY19} were proposed to inject external structured knowledge into the end-to-end model. Among these two, \cite{DBLP:conf/naacl/ZhangSS19} requires converting external knowledge into features while \cite{DBLP:conf/acl/ZhangSSY19} directly uses external knowledge in the format of triples.
    \item \textbf{Stronger Language Representation Models}: Recently, along with the fast development of language representation models, a few works~\cite{DBLP:conf/acl/KantorG19,DBLP:journals/tacl/JoshiCLWZL20} have been trying to replace the encoding layer of the original end-to-end model with more powerful language representation models. Take SpanBERT~\cite{DBLP:journals/tacl/JoshiCLWZL20} as an example, by replacing ELMo with SpanBERT, it boosts the performance by 6.6 F1 over the general coreference resolution task.
\end{enumerate}

\subsection{Performances and Analysis}

\begin{table}[t]
\small
    \centering
    \begin{tabular}{l|c|cc}
        \toprule
      \multirow{ 2}{*}{Model} & \multirow{ 2}{*}{Training data} & \multicolumn{2}{c}{Test data}\\
        && CoNLL & i2b2\\
        \midrule
         \multirow{ 2}{*}{End-to-end} & CoNLL & 72.1  & 75.2  \\
                                  &i2b2 & 20.0  & 92.3  \\
                                  \midrule
        \multirow{ 2}{*}{$\quad$ + KG} &CoNLL & 75.9 & 80.9  \\
                                    &i2b2 & 42.7 & 95.2 \\
        \midrule
        \multirow{ 2}{*}{$\quad$ + SpanBERT} &CoNLL &79.6   &40.8    \\
                                    &i2b2 & 28.5   &80.5   \\

        \bottomrule
    \end{tabular}
    \vspace{0.1in}
    \caption{Models' performance (in F1 score) in cross-domain setting on different training/test data.}
    \label{tab:cross-domain}
\end{table}

We follow the experimental setting of~\cite{DBLP:conf/acl/ZhangSSY19} and test the performance\footnote{We use the released codes of different models along with their default hyper-parameters to finish the experiments. For the end2end model, we also include ELMo~\cite{DBLP:conf/naacl/PetersNIGCLZ18} as part of the representation and thus achieves better performance than the original one in Table~\ref{tab:PCR-vs-generalCR}.} of representative models~\cite{DBLP:conf/emnlp/RaghunathanLRCSJM10,DBLP:conf/acl/ClarkM15,DBLP:conf/emnlp/ClarkM16,DBLP:conf/emnlp/LeeHLZ17,DBLP:conf/acl/ZhangSSY19,DBLP:journals/tacl/JoshiCLWZL20} on the CoNLL-2012 dataset~\cite{DBLP:conf/conll/PradhanMXUZ12}.
From the results in Table~\ref{tab:ordinary-main_result}, we can observe that with the help of the end-to-end model and further modifications, the community has made great progress on the standard evaluation set.
For example, the end-to-end model achieves an F1 score over 70 and adding external knowledge (either in a structured way or a representation way) further boost the performance.
Among all pronoun types, all models perform better on third personal and possessive pronouns, and relatively poorly on demonstrative ones.
This is mainly because of the imbalanced distribution of the dataset (i.e., third personal and possessive pronouns appear much more than demonstrative ones).

\subsubsection{Cross-domain Performance}

To investigate whether current PCR models are good enough to be used in real applications, which could be out of the training domain, we conduct experiments on the cross-domain setting.
In detail, we select two different PCR datasets from different domains (i.e., CoNLL~\cite{DBLP:conf/conll/PradhanMXUZ12} from news and i2b2~\cite{uzuner2012evaluating} from the medical domain) and try to train the model on one dataset and test it on the other.
We conduct experiments with three best-performing models and show the results in Table~\ref{tab:cross-domain}, from which we can see that all models\footnote{SpanBERT performs poorly on i2b2 because the medical corpus is too different from the pre-trained corpus of SpanBERT and we use the default hyper-parameters, which might not be the best ones.} perform significantly worse if they are used across domains.
Compared with the baseline method, adding explicit knowledge can help achieve slightly better performance in the cross-domain setting because its training objective allows models to learn to selectively use suitable knowledge rather than just fitting the training data. 

\subsubsection{Influence of Frequency}
% As shown in , even though teaching models to use external knowledge rather than just fitting the training set can improve the model's performance in the cross-domain setting, the overall performance is still not satisfying.
% This observation indicates that current models are still not good enough to be applied in real applications, where the language usage can be more complex.

    \begin{table}[t]
        \centering
        \small
        \begin{tabular}{l|c|ccc}
            \toprule
            \multicolumn{1}{c|}{Model} & Object Type & P   & R   & F1 \\
            \midrule
            \multirow{2}{*}{End-to-End} & Infrequent & 66.5 & 73.8 & 70.0 \\
            & Frequent & 73.0 & 83.3 & 77.8 \\
            \midrule
            \multirow{2}{*}{\quad + KG} & Infrequent & 77.9 & 72.5 & 75.1\\
            & Frequent & 78.0 & 77.7 & 77.9\\
            \midrule
            \multirow{2}{*}{\quad + SpanBERT} & Infrequent &71.3  &72.4  &71.9  \\ %15877 pairs
            & Frequent &83.3 &85.3 &84.3 \\ % 21363 pairs
            \bottomrule
        \end{tabular}%
        \vspace{0.1in}
        \caption{Influence of the frequency.}
        \label{tab:freq}%
    \end{table}% 

To further analyze the performance of existing models, we split the pronouns based on the frequency of the objects they refer to. If an object appears more than ten times in the whole dataset, we denote it as frequent objects. Otherwise, we denote it as infrequent objects.
As a result, we collect 1,095 frequent and 470,232 infrequent objects, whose average frequencies are 36.2 and 1.46 respectively.
We report the performance of best-performing models on infrequent and frequent objects separately in Table~\ref{tab:freq}.
In general, all models perform better on frequent objects because they appear more in the training data.
Another interesting observation is that even though adding external KG and a stronger language representation model can both boost the performance, their improvements come from different types of objects.
For example, the main contribution of adding KG is on infrequent objects because even though they are less frequent in the training data, they could still be covered by some external knowledge.
As a comparison, using a strong language representation model mainly benefits the frequent objects because it has a stronger ability to fit the training data.
This observation is consistent with our previous observations that adding external KG has more effect on those relatively rare pronouns (i.e., demonstrative pronouns).

% At the same time, we still need to be aware that there are certain doubts behind the current success.
% For example, 75.7 is still not good enough for a PCR model that we can rely on in real life and considering that all models suffer in the cross-domain setting, it is still unclear whether the progress we made in recent years is because we have a strong ability to fit the dataset or the models can better know how to resolve pronouns.
% All of these show that PCR is still a challenging task, which has not been solved yet.
% To help the machine better understand how to resolve pronouns, we need to put more efforts in the future.
% Maybe considering how to better leverage both the training data and external human knowledge and employ a cross-domain setting to evaluate all PCR models would be a good start for the community.

\section{Hard PCR}\label{sec:HardPCR}

\begin{figure}[t]
    \centering
    \includegraphics[width=0.4\linewidth]{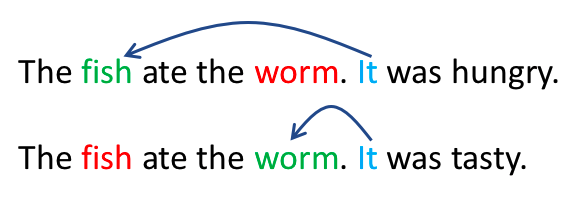}
    \caption{WSC question examples.}
    \label{fig:wino_example}
\end{figure}

As aforementioned, the correct resolution of pronouns requires the inference over both linguistic knowledge and commonsense knowledge.
To clearly reflect how models can resolve pronouns that require the inference over commonsense knowledge, the hard PCR task was proposed.
As Winograd Schema Challenge (WSC) is the most popular hard PCR task, we use the task definition in WSC to define the hard PCR task.
Given a sentence $s$, which contains a pronoun $p$ and two candidates $n_1$, $n_2$, the task is to find out which of the candidates $p$ refers to.
Different from the ordinary PCR task, the influence of all commonly observed features (e.g., gender or plurality) are removed via carefully expert design.
In WSC, all questions are paired up such that questions in each pair have only minor differences (mostly one-word difference), but the answers are reversed.
One pair of the WSC instances is shown in Figure~\ref{fig:wino_example}.
Solving these questions typically requires the support of complex commonsense knowledge.
For example, human beings can know that the pronoun `it' in the first sentence refers to `fish' while the one in the second sentence refers to `worm' because `hungry' is a common property of something eating while `tasty' is a common property of something being eaten. 
Without the support of such commonsense knowledge, answering these questions becomes challenging because both the fish and worm can be hungry or tasty by themselves.

\subsection{Datasets}

We introduce datasets as follows:
\begin{enumerate}[leftmargin=*]
    \item \textbf{Winograd Schema Challenge}: Among all the hard pronoun coreference resolution tasks, WSC is the most popular one. In total, WSC has 273 questions\footnote{The latest version of WSC has 284 questions, but as all the following works are evaluated based on the 273-question version, we still use the 273-question version in this survey.}. Its small size determines that it cannot be used to train a good supervised model and can only be used as the evaluation set.
    \item \textbf{Definite Pronoun Resolution}: Another hard pronoun coreference resolution dataset is the definite pronoun resolution dataset (DPR)\footnote{This dataset is also referred to as WSCR in some works.}~\cite{DBLP:conf/emnlp/RahmanN12}.
    Different from WSC, DPR leveraged undergraduates rather than experts to create the dataset.
    In total, DPR collected 1,886 questions, which is a slightly larger scale than the official WSC.
    However, as DPR could not guarantee that all DPR questions follow the strict design guideline of WSC, questions in DPR are relatively simpler. 
    \item \textbf{WinoGrande}: One common problem of WSC and DPR is their small scales.
    To create a larger scale data, WinoGrande~\cite{DBLP:journals/corr/abs-1907-10641} was proposed.
    By leveraging annotators from Amazon Mechanical Turk, WinoGrande collected 53 thousand WSC-like questions.
    Moreover, to make sure of the dataset quality, WinoGrande applied a bias reduction algorithm to filter out examples that may contain annotation bias.
    Experimental results prove that WinoGrande is much more challenging than the original WSC because the SOTA models on WSC only achieve 51\% accuracy on WinoGrande, which is similar to the random guess.
    \item \textbf{KnowRef}: 
Similar to WinoGrande, KnowRef~\cite{DBLP:conf/acl/EmamiTTSSC19} also aimed at creating a larger scale WSC dataset but with a different approach.
Instead of using crowd-sourcing + adversarial filtering framework, KnowRef tried to extract WSC-like questions from raw sentences.
As a result, KnowRef collected eight thousand WSC-like questions.
\end{enumerate}

\begin{table*}[]
\small
    \centering
    \begin{tabular}{l|l|ccc|cc}
    \toprule
            & Methods & Correct & Wrong & NA & $A_p$ & $A_o$ \\
            \midrule
      \multirow{8}{*}{Unsupervised} &  Random Guess & 137 & 136 & 0 & 50.2\% & 50.2\%  \\
        & Knowledge Hunting~\tiny{\cite{DBLP:conf/emnlp/EmamiCTSC18}} & 119 & 79 & 75 & 60.1\% & 57.3\%  \\
        & SP (Human)~\tiny{\cite{DBLP:conf/acl/ZhangDS19}} & 15 & 0 & 258 & \textbf{100\%} & 52.7\% \\
        & SP (PP)~\tiny{\cite{DBLP:conf/acl/ZhangDS19}} & 50 & 26 & 197 & 65.8\% & 54.4\% \\
        & ASER (String Match)~\tiny{\cite{zhang2019aser}} & 63 & 27 & 183 & 70.0\% & 56.6\% \\
         & LM (Single)~\tiny{\cite{DBLP:journals/corr/abs-1806-02847}} & 149 & 124 & 0 & 54.5\% & 54.5\%  \\
        & LM (Ensemble)~\tiny{\cite{DBLP:journals/corr/abs-1806-02847}} & 168 & 105 & 0 & 61.5\% & 61.5\% \\
        & GPT-2~\tiny{\cite{radford2019language}} & 193 & 80 & 0 & 70.7\% & 70.7\% \\
    \midrule
       \multirow{6}{*}{Finetuning} 
       & BERT~\tiny{\cite{DBLP:conf/naacl/DevlinCLT19}}  \small{+ASER}~\tiny{\cite{zhang2019aser}} & 177 & 96 & 0 & 64.5\% & 64.5\% \\
       & BERT~\tiny{\cite{DBLP:conf/naacl/DevlinCLT19}}  \small{+DPR}~\tiny{\cite{DBLP:conf/emnlp/RahmanN12}} & 195 & 78 & 0 & 71.4\% & 71.4\% \\
    %   & BERT~\cite{joshi-etal-2019-bert} + DPR~\cite{DBLP:conf/emnlp/RahmanN12} + ASER~\cite{zhang2019aser} & 200 & 73 & 0 & 73.3\% & 73.3\% \\
       & BERT~\tiny{\cite{DBLP:conf/naacl/DevlinCLT19}}  \small{+WinoGrande}~\tiny{\cite{DBLP:journals/corr/abs-1907-10641}} & 210 & 63 & 0 & 76.9\% & 76.9\% \\
       & RoBERTa~\tiny{\cite{DBLP:journals/corr/abs-1907-11692}}  \small{+DRP}~\tiny{\cite{DBLP:conf/emnlp/RahmanN12}} & 227 & 46 & 0 & 83.1\% & 83.1\% \\
       & RoBERTa~\tiny{\cite{DBLP:journals/corr/abs-1907-11692}}  \small{+WinoGrande}~\tiny{\cite{DBLP:journals/corr/abs-1907-10641}} & 246 & 27 & 0 & 90.1\% & \textbf{90.1\%} \\
       \midrule
       \multirow{2}{*}{Human Beings}
       & Original~\tiny{\cite{levesque2012winograd}} & 252 & 21 & 0 & 92.1\% & 92.1\% \\
        & Recent~\tiny{\cite{DBLP:journals/corr/abs-1907-10641}} & 264 & 9 & 0 & 96.5\% & 96.5\% \\
       \bottomrule
    \end{tabular}
    \vspace{0.1in}
    \caption{Performances of different models on the 273-question version WSC. $NA$ means that the model cannot give a prediction, $A_p$ means the accuracy of predict examples without $NA$ examples, and $A_o$ the overall accuracy.}
    \label{tab:WSC}
\end{table*}

\subsection{Methods}

In this subsection, we introduce existing approaches for the hard PCR task. 
As the majority of the methods are evaluated based on WSC, all the discussion and analysis are based on their performance on WSC.

\subsubsection{Reasoning with Structured Knowledge}

At first, people tried to leverage different commonsense knowledge resources to solve WSC questions in an explainable way. For example, \cite{liu2016commonsense} first leveraged the commonsense triplets from ConceptNet~\cite{liu2004conceptnet} to train the word embeddings and then applied the embeddings to solve the WSC task. Knowledge hunter~\cite{DBLP:conf/emnlp/EmamiCTSC18} proposed to leverage search engines (e.g., Google) to acquire needed commonsense knowledge. It first searched WSC questions in search engines and then used the returned searching results to solve WSC questions. SP-10K~\cite{DBLP:conf/acl/ZhangDS19} conducted experiment to show that selectional preference (SP) knowledge such as human beings are more likely to eat `food' rather than `rock' can also be helpful for solving WSC questions. Last but not least, ASER~\cite{zhang2019aser} tried to use knowledge about eventualities (e.g., `being hungry' can cause `eat food') to solve WSC questions. In general, structured commonsense knowledge can help solve one-third of the WSC questions, but their overall performance is limited due to their low coverage.
There are mainly two reasons: (1) coverage of existing commonsense resources are not large enough; (2) lack of principle way of using structured knowledge for NLP tasks. Current methods~\cite{DBLP:conf/emnlp/EmamiCTSC18,DBLP:conf/acl/ZhangDS19,zhang2019aser} mostly rely on string match. However, for many WSC questions, it is hard to find supportive knowledge in such way.

\subsubsection{Language Representation Models}

Another approach is leveraging language models to solve WSC questions~\cite{DBLP:journals/corr/abs-1806-02847}, where each WSC question is first converted into two sentences by replacing the target pronoun with the two candidates respectively and then the language models can be employed to compute the probability of both sentences.
The sentence with a higher probability will be selected as the final prediction.
As this method does not require any string match, it can make prediction for all WSC questions and achieve better overall performance.
Recently, a more advanced transformer-based language model GPT-2~\cite{radford2019language} achieved better performance due to its stronger language representation ability.
The success of language models demonstrates that rich commonsense knowledge can be indeed encoded within language models implicitly.
Another interesting finding about these language model based approaches is that they proposed two settings to predict the probability: (1) Full: use the probability of the whole sentence as the final prediction; (2) Partial: only consider the probability of the partial sentence after the target pronoun. 
Experiments show that the partial model always outperforms the full model.
One explanation is that the influence of the imbalanced distribution of candidate words is relieved by only considering the sentence probability after them. Such observation also explains why GPT-2 can outperform unsupervised BERT on WSC because models based on BERT, which relies on predicting the probability of candidate words, cannot get rid of such noise.

\subsubsection{Fine-tuning  Representation Models}

Last but not least, we would like to introduce current best-performing models on the WSC task, which fine-tunes pre-trained language representation models (e.g., BERT~\cite{DBLP:conf/naacl/DevlinCLT19} or RoBERTa~\cite{DBLP:journals/corr/abs-1907-11692}) with a similar dataset (e.g., DPR~\cite{DBLP:conf/emnlp/RahmanN12} or WinoGrande~\cite{DBLP:journals/corr/abs-1907-10641}).
This idea was originally proposed by \cite{DBLP:conf/acl/KocijanCCYL19}, which first converts the original WSC task into a token prediction task and then selects the candidate with higher probability as the final prediction.
In general, the stronger the language model and the larger the fine-tuning datasets are, the better the model can perform on the WSC task.

\subsection{Performances and Analysis}

To clearly understand the progress we have made on solving hard PCR problems, we show the performance of all models on Winograd Schema challenge in Table~\ref{tab:WSC}. From the results, we can make the following observations:
\begin{enumerate}[leftmargin=*]
    \item Even though methods that leverage structured knowledge can provide explainable solutions to WSC questions, their performance is typically limited due to their low coverage. 
    \item Different from them, language model based methods represent knowledge contained in human language with an implicit approach, and thus do not have the matching issue and achieve better overall performance.
    \item In general, fine-tuning pre-trained language representation models (e.g., BERT and RoBERTa) with similar datasets (e.g., DPR and WinoGrande) achieve the current SOTA performances and two observations can be made: 
    (1) The stronger the pre-trained model, the better the performance. This observation shows that current language representation models can indeed cover commonsense knowledge and along with the increase of their representation ability (e.g., deeper model or larger pre-training corpus like RoBERTa), more commonsense knowledge can be effectively represented.
    (2) The larger the fine-tuning dataset, the better the performance. This is probably because the knowledge about some WSC questions is only covered by Winogrande but not in DPR.
\end{enumerate}

% \subsubsection{Further Discussion}

\begin{table}[t]
\small
    \centering
    \begin{tabular}{c||c|c|c||c}
    \toprule
    \multirow{2}{*}{\diagbox{L.R.}{Rel.}} & High & Medium & Low & Overall \\
     & (13,466) & (13,466) & (13,466) & (40,398)\\
    \midrule 
    1e-6      & 87.81\%& 85.63\%& 84.95\%& \textbf{88.89\%} \\
    2e-6    & 87.46\% & 87.81\% & 50.53\% & \textbf{90.32\%}\\
    5e-6      & 87.10\%& 86.74\%& 50.17\% & \textbf{91.76\%} \\
    \midrule
    % 1e-5 (default)     & 86.74\%& 88.17\%& 86.02\%& 87.46\% \\
    1e-5 (default)     & \textbf{87.81\%}& 85.66\%& 84.94\%& 87.46\% \\
    \midrule
    2e-5      & 53.04\%& 51.25\%& 52.33\%& \textbf{84.58\%} \\
    5e-5 & 51.97\% & 50.09\% & 51.97\% & \textbf{55.56\%}\\
    1e-4 & \textbf{53.75\%} & 53.05\% & 52.69\% & 51.06\%\\
    \midrule
    Average & 72.71\% & 71.46\% & 61.08\% & \textbf{78.51\%}\\
    
    % (5e-6,1e-5,2e-5) - (xx,84.23,xx);()
    \bottomrule
    \end{tabular}
    \vspace{0.1in}
    \caption{Performance of fine-tuning RoBERTa with different subsets of Winogrande and different learning rates. L.R. means learning rate and Rel. means relevance to WSC data. WinoGrande instances are grouped into three subsets. Numbers of instances are shown in brackets. Best performed datasets for each learning rate is indicated with the \textbf{bold} font.}
    \label{tab:detailed_result}
\end{table}

To investigate the reason behind WinoGrande's success, we divide WinoGrande into subsets based on the instances' relevance towards WSC. Assume that the instance set of WinoGrande and WSC are $\IM_{WG}$ and $\IM_{WSC}$ respectively, for each instance $i \in \IM_{WG}$, we design its relevance score as follows:
\begin{equation}
    R_{WSC}(i) = Max (\frac{O^2(i, i^\prime)}{L(i) \cdot L(i^\prime)}, i^\prime \in \IM_{WSC}),
\end{equation}
where $O(i, i^\prime)$ is the unigram co-occurrence of $i$ and $i^\prime$ and $L()$ the instance length. We use the released code and dataset to conduct the experiments and follow all hyper-parameters as the original paper~\cite{DBLP:journals/corr/abs-1907-10641} except the batch size\footnote{The original batch size is 16 and our batch size is 4 due to the GPU memory limitation, so the experimental result is slightly different from the one reported in the original paper.}.

From the results in Table~\ref{tab:detailed_result}, we can observe that: (1) The most relevant instances contribute the most to the success. In some learning rate settings, it performs similar to or even better than the overall set; (2) Less relevant instances also help, which shows that current fine-tuning approach is not just fitting the data but also learning some underneath knowledge about solving the task from the data; (3) The model can be sensitive to the hyper-parameters (i.e., learning rate). 
Different subsets have different best hyper-parameters and the learning process can easily fail with a bad hyper-parameter. 
To achieve a good performance on a fixed dataset like WSC, we can tune the hyper-parameters. But to create a reliable PCR system we can rely on in real life, we probably need a more robust model.
\section{Other PCR Tasks}\label{sec:OtherPCR}

Besides the ordinary and hard PCR tasks, PCR is also an important research topic for many special purposes (e.g., gender bias) or in some special settings (e.g., Visual-aware PCR). 
In this section, we briefly introduce these tasks:

\begin{enumerate}[leftmargin=*]
    \item \textbf{PCR in the Medical Domain}: I2b2~\cite{uzuner2012evaluating} is a dataset that focuses on identifying coreference relations in electrical medical records. 
As reported in \cite{DBLP:conf/acl/ZhangSSY19}, the training set of I2b2 contains 2,024 third personal pronouns, 685 possessive pronouns, and 270 demonstrative pronouns. Its test set contains 1,244 third personal pronouns, 367 possessive pronouns, and 166 demonstrative pronouns.
As a dataset in a relatively narrow domain, the usage of domain knowledge becomes important.
As shown in \cite{DBLP:conf/acl/ZhangSSY19}, i2b2 can be used as an additional dataset to evaluate models' cross-domain abilities.

\item \textbf{PCR for Machine Translation}: ParCor~\cite{DBLP:conf/lrec/GuillouHSTW14} and ParCorFull~\cite{lapshinova-koltunski-etal-2018-parcorfull} are datasets focusing on PCR in parallel multi-lingual datasets, which can be used in downstream machine translation tasks.
Different from other PCR works, it focuses on how to leverage the PCR results for better translation rather than how to solve the PCR problem.

\item \textbf{PCR for Chatbots}: CIC~\cite{DBLP:conf/sigdial/ChenC16} is a dataset focusing on identifying coreference relations in multi-party conversations. Compared with the ordinary PCR tasks, which are mostly annotated on formal textual data (e.g., newswire), identifying coreference relation in conversation is more challenging. 

\item \textbf{PCR for Studying Gender Bias}: Nowadays, gender bias has been a hot research topic in the NLP community~\cite{DBLP:conf/naacl/RudingerNLD18,DBLP:conf/naacl/ZhaoWYOC18}. Among all the works, WinoGender~\cite{DBLP:conf/naacl/RudingerNLD18} is one of the most popular ones.
The setting of WinoGender is similar to the setting of WSC~\cite{levesque2012winograd}, where each sentence contains one target pronoun and two candidate noun phrases and the models are required to select the correct antecedent from the two candidates.
But the purpose is different.
WSC aims at evaluating models' abilities to understand commonsense knowledge, while WinoGender aims at evaluating how well models can predict without the influence of gender bias.
The experiments show that some gender bias (e.g., `he' is more likely to be predicted to be the doctor rather than the nurse by the machine) indeed exists in pre-trained language representation models. 
Such observation is astonishing and motivates the community to think about how to minimize the influence of such gender bias.

\item \textbf{Visual-aware PCR}: Recently, a visual-aware PCR dataset~\cite{DBLP:conf/emnlp/YuZSSZ19}, which evaluates how well models can ground pronouns to visual objects, was proposed. 
Similar to CIC~\cite{DBLP:conf/sigdial/ChenC16}, Visual-PCR also focuses on pronouns in daily dialogue, where the language usage is informal and a lot of background knowledge could be missing.
For example, if one speaker refers to something both speakers can see, they may directly use a pronoun rather than introduce it first.
In such a case, a pronoun may refer to not mentioned objects in the conversation.
As analyzed in the original paper, 15\% of pronouns in conversations refer to not mentioned objects and for them, leveraging the visual context information becomes crucial.
As shown in~\cite{DBLP:conf/eccv/KotturMPBR18}, grounding pronouns to the visual objects can significantly help the model to better understand the dialog and generate the better response, which further proves that visual PCR is an important research topic worth exploring.
\end{enumerate}

\section{Conclusion}
\label{sec:conclusion}

In this paper, we survey the progress on the pronoun coreference resolution (PCR) task and the limitation of existing approaches.
Experiments and analysis on both the ordinary and hard PCR tasks demonstrate that even though we have made great progress based on the main evaluation metric, the PCR task is still far away from being solved.
For example, all best-performing ordinary PCR models struggle on the cross-domain setting as well as infrequent objects, and even though fine-tuning pre-trained language representation models can achieve near-human performance on WSC, it can be sensitive to the hyper-parameters.
All codes will be released to encourage the research on the PCR task.

\section*{Acknowledgements}
This paper was supported by  Early Career Scheme (ECS, No. 26206717), General Research Fund (GRF, No. 16211520), and Research Impact Fund (RIF, No. R6020-19) from the Research Grants Council (RGC) of Hong Kong.

\bibliographystyle{unsrt}  
\bibliography{references}  %%% Remove comment to use the external .bib file (using bibtex).


% George Kour and Raid Saabne.
% \newblock Real-time segmentation of on-line handwritten arabic script.
% \newblock In {\em Frontiers in Handwriting Recognition (ICFHR), 2014 14th
%   International Conference on}, pages 417--422. IEEE, 2014.

% George Kour and Raid Saabne.
% \newblock Fast classification of handwritten on-line arabic characters.
% \newblock In {\em Soft Computing and Pattern Recognition (SoCPaR), 2014 6th
%   International Conference of}, pages 312--318. IEEE, 2014.

% \bibitem{hadash2018estimate}
% Guy Hadash, Einat Kermany, Boaz Carmeli, Ofer Lavi, George Kour, and Alon
% \newblock Estimate and replace: A novel approach to integrating deep neural
%   networks with existing applications.
% \newblock {\em arXiv preprint arXiv:1804.09028}, 2018.

% \end{thebibliography}

\end{document}